\documentclass[runningheads]{llncs}
\usepackage{graphicx}
\usepackage{mathptmx} 
\usepackage{amsmath, amsfonts, amssymb} 
\usepackage{helvet, courier} 
\usepackage{microtype} 
\usepackage[colorlinks,linkcolor=blue]{hyperref} 
\usepackage{caption, subcaption} 
\usepackage{booktabs} 
\usepackage{nicefrac} 
\usepackage{color, gensymb} 
\usepackage{url} 
\usepackage[utf8]{inputenc} 
\usepackage{verbatim} 
\usepackage{placeins} 
\usepackage{tabularx}
\usepackage{cite}
\usepackage{multirow} 
\usepackage{amssymb}
\usepackage{booktabs}
\usepackage{mathptmx} 
\usepackage{bm}
\begin{document}
\title{MVL-Loc: Leveraging Vision-Language Model for Generalizable Multi-Scene Camera Relocalization
}  

%
%
\author{Zhendong Xiao\inst{1} \and
Wu Wei\inst{1} \and
Shujie Ji\inst{1} \and Shan Yang\inst{1} \and Changhao Chen\inst{2} }
\authorrunning{Z. Xiao et al.}
%
\institute{School of Automation Science and Engineering, South China University of Technology, Guangzhou, Guangdong Province, China \and
Thrust of Intelligent Transportation and Thrust of Artificial Intelligence, The Hong Kong University of Science and Technology (Guangzhou), Guangzhou, China\thanks{Corresponding author: Changhao Chen, changhaochen@hkust-gz.edu.cn}
\email{ auxiao2022@mail.scut.edu.cn}
}
%
\maketitle
%

\begin{abstract}
Camera relocalization, a cornerstone capability of modern computer vision, accurately determines a camera's position and orientation (6-DoF) from images and is essential for applications in augmented reality (AR), mixed reality (MR), autonomous driving, delivery drones, and robotic navigation.
Unlike traditional deep learning-based methods regress camera pose from images in a single scene which lack generalization and robustness in diverse environments. We propose MVL-Loc, a novel end-to-end multi-scene 6-DoF camera relocalization framework. MVL-Loc leverages pretrained world knowledge from vision-language models (VLMs) and incorporates multimodal data to generalize across both indoor and outdoor settings. Furthermore, natural language is employed as a directive tool to guide the multi-scene learning process, facilitating semantic understanding of complex scenes and capturing spatial relationships among objects. Extensive experiments on the 7Scenes and Cambridge Landmarks datasets demonstrate MVL-Loc’s robustness and state-of-the-art performance in real-world multi-scene camera relocalization, with improved accuracy in both positional and orientational estimates.

\end{abstract}

\section{INTRODUCTION}

\begin{figure}[t]
    \centering
    \includegraphics[width=0.95\columnwidth]{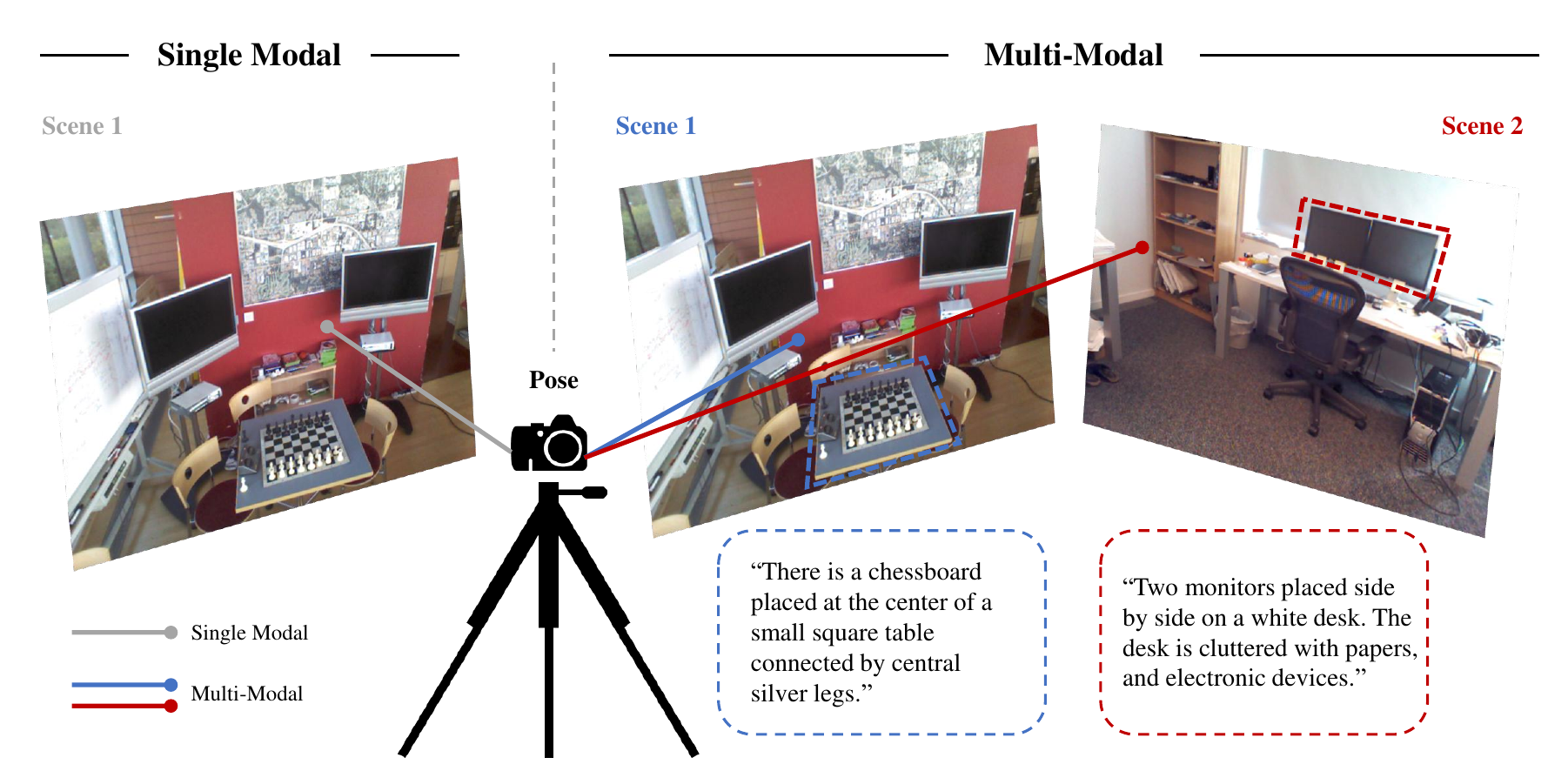}
    \captionsetup{justification=justified}
    
    \caption{Language guides multi-scene learning for camera pose estimation, even when visual appearances vary. Our MVL-Loc uses contextual cues, such as a chessboard on a small table (Scene 1) and dual monitors on a cluttered desk (Scene 2), to estimate the camera's pose.
}
    \label{fig1}
\end{figure}

Camera relocalization, which estimates a camera’s 6-DoF pose from visual inputs, is a fundamental problem in intelligent systems, enabling applications from autonomous navigation to augmented reality (AR).Traditional approaches solve this task by matching visual features to pre-constructed maps and estimating poses via algorithms such as PnP and Kabsch, combined with RANSAC \cite{fischler1981random}, to establish 2D-3D correspondences. Although effective, these methods are computationally intensive and storage-heavy, as they rely heavily on the quantity and quality of reference data.
Deep learning-based camera relocalization redefines the paradigm by enabling end-to-end pose regression directly from raw pixels, bypassing the need for handcrafted features. PoseNet\cite{kendall2015posenet} pioneered this approach, estimating 6-DoF poses from single images using GoogLeNet. Subsequent improvements include LSTM modules\cite{walch2017image}, ResNet34 skip connections\cite{melekhov2017image}, uncertainty modeling\cite{kendall2016modelling}, and geometric constraints\cite{kendall2017geometric}. Video-based methods such as VidLoc~\cite{clark2017vidloc} leverage temporal constraints, while AtLoc\cite{wang2020atloc} enhances pose consistency with attention mechanisms. EffLoc\cite{xiao2024effloc} optimizes efficiency using a Vision Transformer with memory-efficient self-attention. Despite these advances, deep learning methods remain inherently limited in cross-scene generalization, struggling to adapt to unseen layouts due to scene-specific training dependency.

To transcend scene-specific constraints, recent advances in embodied AI have integrated vision-language models (VLMs) to augment transferability with cross-modal semantic reasoning. Enhanced with pre-trained web knowledge, these models have demonstrated significant progress. NavGPT \cite{zhou2023navgpt} utilizes VLMs to interpret and execute complex navigational commands in natural language, integrating these instructions with real-time visual inputs to navigate diverse environments effectively. Similarly, RT-2 \cite{rt22023arxiv} merges visual perception with natural language processing, enabling service robots to respond with heightened context awareness. Nevertheless, no prior work has systematically leveraged VLMs for joint semantic-geometric optimization in camera relocalization.

In this work, we propose \textbf{MVL-Loc}, a novel \textbf{M}ulti-scene \textbf{V}isual \textbf{L}anguage \textbf{Loc}alization framework, which incorporates a Vision-Language Model trained across multiple scenes for 6-DoF camera relocalization. As illustrated in Figure \ref{fig1}, we assign each scene unique, non-template-based, language-guided instructions to enhance the fusion of multi-scene settings. These specific contextual cues enable the model to differentiate between scenes and establish robust spatial and semantic relationships among objects.  Our model surpasses existing multi-scene approaches, demonstrating superior accuracy in  image-based camera relocalization across indoor and outdoor benchmarks. Specifically, MVL-Loc surpasses MSPN \cite{blanton2020extending} on the 7-Scenes dataset, reducing the average position error by 23.8\% and rotation error by 19.2\%. On the Cambridge Landmarks dataset, MVL-Loc improves position error by 6\% and rotation error by 7.3\% compared to C2f-MS-Transformer \cite{shavit2023c2f}, a recent transformer-based multi-scene approach.
In summary, our main contributions are as follows:
\begin{itemize}

\item  We propose MVL-Loc, a novel multi-scene camera relocalization framework that harnesses pretrained world knowledge from vision-language models (VLMs), effectively generalizing to both indoor and outdoor environments.
\item  We leverage natural language as a guiding tool for the multi-scene learning process of VLMs, enabling a deep semantic understanding of complex scenes and capturing the spatial relationships between objects and scenes.
\item Through extensive experiments on common benchmarks, i.e. 7-Scenes and Cambridge Landmarks datasets, we demonstrate that our MVL-Loc framework achieves state-of-the-art performance in the task of end-to-end multi-scene camera relocalization.
\end{itemize}

\section{Related Work}

\subsection{Deep Learning for Camera Relocalization}

Camera relocalization aims to recover the 6-DoF pose of a camera given an input image. Traditional approaches rely on geometric pipelines involving feature matching, structure-from-motion, and PnP with RANSAC~\cite{chen2011city, shen2025imagdressing, shen2024imagpose}. However, these methods are sensitive to viewpoint changes and require extensive 3D maps.
Recent advances in deep learning have enabled end-to-end pose regression directly from images. PoseNet~\cite{kendall2015posenet} initiates this line by predicting camera pose without requiring explicit 2D-3D correspondences. Subsequent methods integrate temporal cues via RNNs~\cite{walch2017image} or LSTMs, as in VidLoc~\cite{clark2017vidloc}, to improve stability in video streams. PoseNetV2~\cite{kendall2017geometric} introduces geometric reprojection loss for spatial consistency, while DSAC~\cite{brachmann2017dsac} and HybridPose~\cite{camposeco2019hybrid} revisit differentiable RANSAC for learning-based correspondence selection.
Map-free~\cite{arnold2022mapfree} employs depth-guided relative pose loss to remove map dependencies. AtLoc~\cite{wang2020atloc} and EffLoc~\cite{xiao2024effloc} further enhance localization accuracy via attention mechanisms and efficient vision transformers, respectively.
Generalization across diverse environments remains challenging. Multi-Scene PoseNet~\cite{blanton2020extending} attempts to train shared pose regressors, while transformer-based models like Ms-Transformer~\cite{shavit2021mstrans} and its coarse-to-fine variant C2f-MS-Transformer~\cite{shavit2023c2f} offer better scalability. MapNet~\cite{brahmbhatt2018geometry} fuses visual odometry and GPS, and FusionLoc~\cite{lee2023fusionloc} incorporates LiDAR to augment visual features.

\subsection{Vision-Language Models for Geospatial Reasoning}

The emergence of large vision-language models (LVLMs) has reshaped cross-modal learning. CLIP~\cite{radford2021learning} demonstrates strong generalization by aligning vision and language in a contrastive manner, inspiring downstream tasks such as image editing~\cite{gao2021clip, li2022envedit}, scene graph grounding~\cite{zhong2021SGGfromNLS}, and 3D scene understanding~\cite{rao2021denseclip, Mirjalili2023fmloc}.
Particularly relevant is CLIP-Loc~\cite{Matsuzaki2024CLIPLocML}, which utilizes CLIP to associate object-level semantics with visual correspondences for localization. However, its dependence on PROSAC-based ranking limits adaptability to dynamic scenes. In contrast, our method eliminates explicit matching through end-to-end learning, guided by language embeddings that encode spatial priors and object semantics.
Recent studies extend LVLMs to geospatial domains. GeoLLM~\cite{manvi2024geollm} leverages LLMs to infer spatial distributions, and GeoReasoner~\cite{li2024georeasoner} performs coarse-grained city-level localization from street-view imagery. Yet these methods operate at low spatial resolution and fail in precise tasks such as camera pose estimation.

Our approach draws insights from semantic modeling in generative tasks like progressive pose-conditioned generation~\cite{shen2023advancing} and rich-contextual diffusion frameworks for spatial consistency~\cite{shen2025boosting}. We adapt these ideas to pose estimation, enabling fine-grained semantic-to-spatial mapping through pre-trained vision-language priors.

\section{Multi-Scene Visual Language Localization}

\begin{figure*}[t]
    \centering
    \includegraphics[width=\textwidth,height=0.3\textheight]{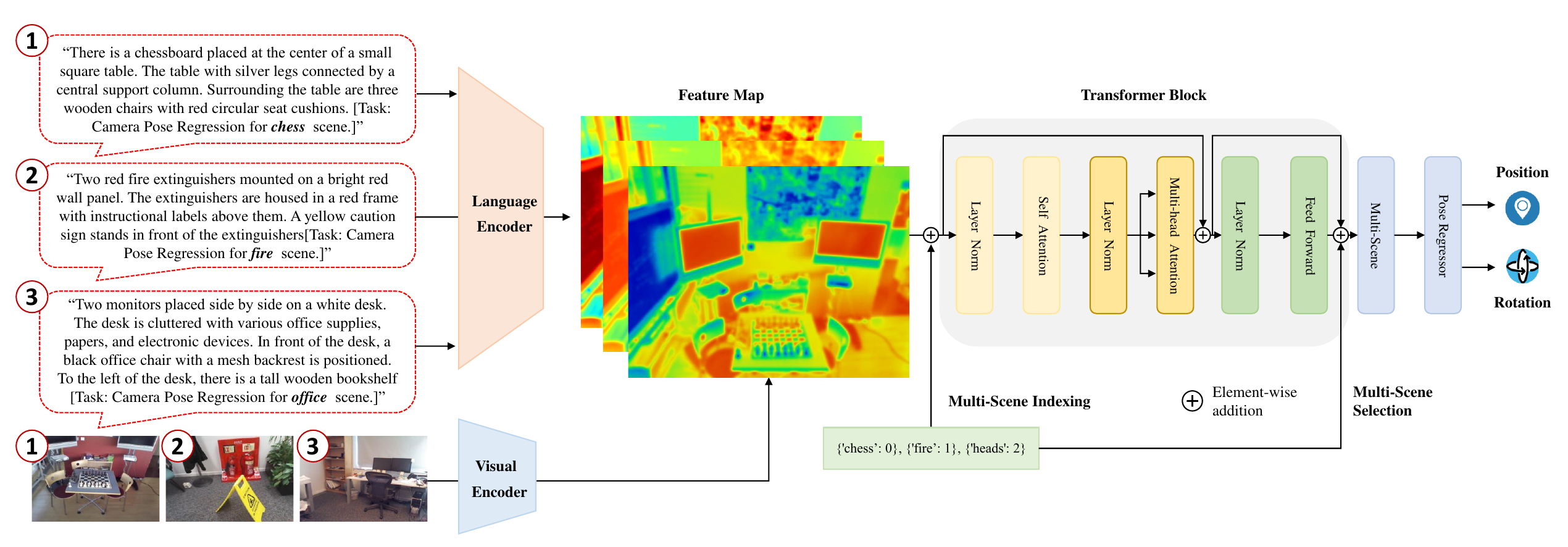}
    \caption{Overview of the MVL-Loc framework for multi-scene camera relocalization using open-world language guidance. The left section shows natural language descriptions guiding the model across three scenes. The center section depicts multi-modal input processing, integrating visual-language correspondence via Transformer blocks with self-attention and multi-head attention. The right section highlights the final 6-DOF camera pose output through multi-scene selection and pose regression.}
    \label{fig:framework}
\end{figure*}

Language-guided visual localization remains underexplored for multi-scene camera relocalization. We present MVL-Loc, a framework that regresses 6-DoF camera poses from single images across multiple scenes using open-set natural language descriptions as contextual guidance. By jointly processing visual inputs and language embeddings, the model achieves cross-scene generalization without requiring scene-specific fine-tuning.

\subsection{MVL-Loc Framework}
The pipeline of our framework is shown in Figure \ref{fig:framework}, which consists of Language-guided text-to-scene correspondences generation, and multi-scene camera pose estimation. In text-to-scene correspondences generation, the vision-language integration is based on the similarity of the description embeddings and vision feature embeddings. The localized region features \( \mathring{V} \in \mathbb{R}^{C \times H \times W} \)  are extracted via 2D convolutional encoder from the image \( I \in \mathbb{R}^{C \times H \times W} \) to construct spatial hierarchies:
\begin{equation}
    \mathring{V} = f_{\text{conv}}(I) + \gamma \cdot P(I),
\end{equation}
Where \( \gamma \) is the negative square root of the width $W$, which ensures the positional embedding \( P \in \mathbb{R}^{C \times H \times W} \) does not overshadow the visual input.
And the descriptions text $T$ can vary in length across different scenes, but they do describe subsets of the objects related to the scenes.
\begin{equation}
    \mathring{L} = f_{\text{token}}(T) + \gamma \cdot P(T).
\end{equation}
For the Transformer Block, we  employed \( N \) identical standard Decoder layers to classify $K$ scenes and experimented with \( N = 2, 4, 8, \) and 16. The model failed to converge when \( N = 2 \), and at \( N = 8 \) and 16, the computational demand was prohibitively high. Each of the four layers comprises self-attention (SA) and multi-head attention (MHA). In the model, each layer $n$, $n=1..N$ applies layer normalization before each module and incorporates residual connections, as detailed in the following formula:
\begin{equation}
\mathbf{({V}, {L})}^{n^{\prime }}=SA(LN(\mathbf{({V}, {L})}^{n-1}))+\mathbf{({V}, {L})}^{n-1}\in \mathbb{R}^{C \times H \times W}, \label{eq:decoder-sa}
\end{equation}
\begin{equation}
\mathbf{({V}, {L})}^{n}=MHA(LN(\mathbf{({V}, {L})}^{n^{\prime }}))+\mathbf{({V}, {L})}^{n^{\prime }}\in \mathbb{R}^{C \times H \times W}.    \label{eq:decoder-mha}
\end{equation}

At the final layer \( N \), the output is processed through a feedforward module with GELU nonlinear activation:
\begin{equation}
\mathbf{({V}, {L})}^{N}=FF(LN(\mathbf{({V}, {L})}^{n}))+\mathbf{({V}, {L})}^{n}\in \mathbb{R}^{C \times H \times W}    \label{eq:decoder-ff}
\end{equation}

\begin{figure}[t]
  \centering
  \begin{subfigure}[b]{0.31\columnwidth}
    \centering
    \includegraphics[width=\linewidth,height=0.7\columnwidth]{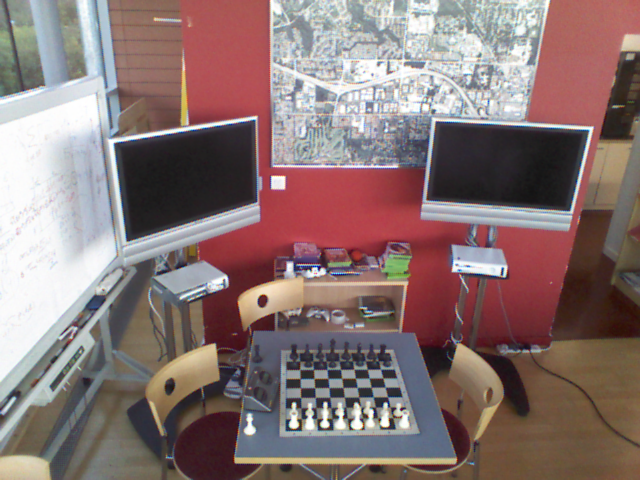}
    \subcaption{Original image}
    \label{fig4-0}
  \end{subfigure}%
  \hspace{0.015\columnwidth}
  \begin{subfigure}[b]{0.31\columnwidth}
    \centering
    \includegraphics[width=\linewidth,height=0.7\columnwidth]{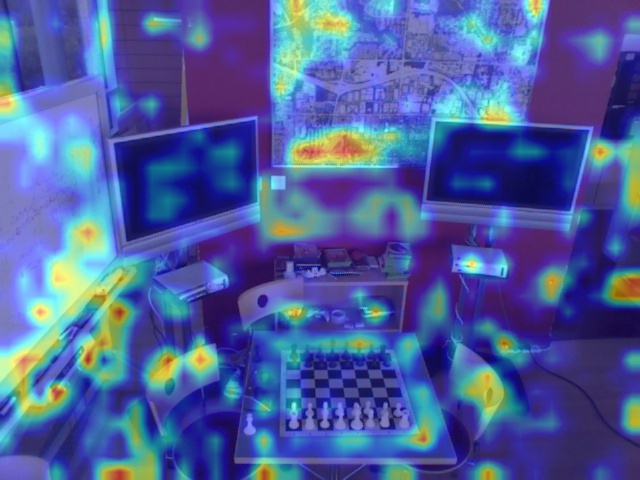}
    \subcaption{Broad language-guided}
    \label{fig4-1}
  \end{subfigure}%
  \hspace{0.015\columnwidth}
  \begin{subfigure}[b]{0.31\columnwidth}
    \centering
    \includegraphics[width=\linewidth,height=0.7\columnwidth]{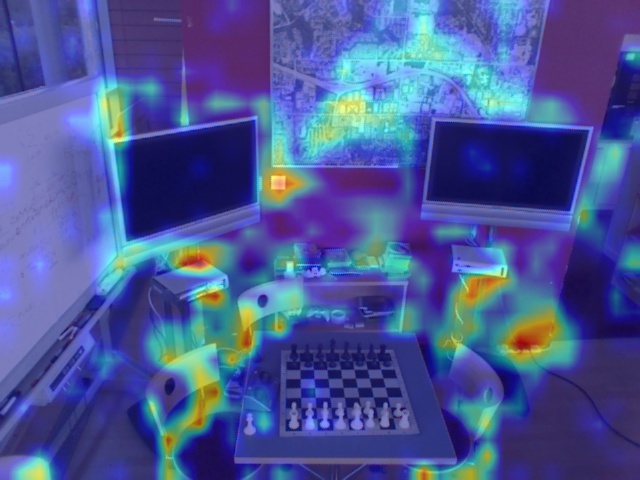}
    \subcaption{Detailed language-guided}
    \label{fig4-2}
  \end{subfigure}
  \caption{Attention visualizations under the \textit{Chess} scene. From left to right: original image, broad language-guided attention, and detailed language-guided attention with sharper focus on the chessboard surrounding objects.}
  \label{fig:chess_att}
\end{figure}

\subsection{Language-Guided Camera Relocalization}
We propose a cross-modal fusion encoder (Enc) to integrate natural language descriptions with visual features for enhanced multi-scene camera pose estimation. These descriptions, which emphasize spatial relationships (e.g., "in front of," "surrounding"), are fused with visual features via dot product:
\begin{equation}
\mathbf{({V}, {L})} = \text{Enc}[\mathring{V} \cdot \mathring{L}^{\text{T}}], \label{eq:VLfusion}
\end{equation}

where \( (\mathring{V}, \mathring{L}) \) and \( (\mathbf{V}, \mathbf{L}) \) represent the vision and language features before and after fusion, respectively. Furthermore, we use the transpose of the language to  align well with the visual features.

Unlike CLIP’s fixed templates (e.g., "a photo of a [scene]"), we introduce dynamic, learnable language descriptions tailored to camera relocalization, focusing on object spatial relationships. Detailed descriptions, like "A chessboard on a small table surrounded by chairs," direct the model's attention to key elements, improving pose estimation by aligning spatial and semantic cues. Broad descriptions, such as "A chessboard in a classroom," disperse the model's focus as shown in Figure~\ref{fig:chess_att}, reducing precision. Moreover, comprehensive prompts such as "Two monitors side by side on a cluttered desk with a chair in front," help the model localize key scene elements (monitors, desk, chair) more accurately. This approach leverages spatial and commonsense knowledge, enhancing relocalization in complex environments. MVL-Loc aligns language and visual data to achieve more precise and context-aware camera relocalization.

\subsection{Multi-Scene Camera Pose Regression}
Based on the single-scene language-guided training, we extend this methodology by indexing each scene and pairing it with its corresponding detailed description.
Unlike Multi-Scene PoseNet (MSPN)\cite{blanton2020extending}, which produces separate models for each corresponding scene. We train different scenes in parallel for all poses, where each scene is guided by its unique set of language prompts. The final decoder layer outputs the pose sequences corresponding to each scene's latent features. Since a single image corresponds to the specific scene from which it was captured, the appropriate latent embedding must be selected.
We index the Scene list \( {K}= [1, 2, \dots, k]\) represents the number of predicted scene and pair with its corresponding detailed description. The input list is {[``Chess", ``Fire", ``Heads", \dots,``Red kitchen"] or [``King's College", ``Old hospital", \dots,``Stmary's church"]}.
In order to determine each query image corresponds to the scene from which the image was taken, we pass the classify logits \( z = [z_1, z_2, \dots, z_k] \) through the SoftMax function to obtain  predicted probabilities:
\begin{equation}
\pi_{i} = \frac{\exp(k_i)}{\sum_{j=1}^{K} \exp(k_j)} \quad \text{for } i = 1, 2, \dots, k,
\end{equation}

where \( \pi_{i} \) is the probability that the input image belongs to the \( i \)-th scene.
Our work builds upon the VidLoc\cite{brahmbhatt2018geometry} pose estimation method, which regresses 6-DoF camera poses from vision-language fused features through Multilayer Perceptrons (MLPs) corresponding to the selected scene \(i\) with the maximal probability \( \pi_{i} \):
\begin{equation}
\left[p,q\right]_{{i}} =MLPs(\mathbf{({V}, {L})}),
\end{equation}
here $p \in \mathbb{R}^{3}$ represented the 3D camera position, and $q \in \mathbb{R}^{4}$ represented a 4D unit quaternion for orientation.
We first compute the  Negative Log Likelihood loss for classification, given the ground truth scene index ${k}_{0}$:
\begin{equation}
L_{cls} = -\log(\pi_{i}). 
\end{equation}
We then integrate the camera pose loss and multi-scene classification loss to get the overall loss:
\begin{equation}
L_{\mathbf{ms}}=|p-\hat{p}|_1e^{-\alpha}+\alpha+|\log{q}-\log{\hat{q}}|_1e^{-\beta}+\beta+ L_{cls},
\end{equation}
$\alpha$ and $\beta$ are the scale factors to balance the positional and orientational losses. We implemented the logarithm of a unit quaternion instead of unit-norm quaternion, which provides a minimally parameterized representation in three dimensions. Hence, it allows for L1 distance loss without requiring normalization. The L1 loss mitigates the influence of outliers, enhancing robustness to unusual observations while encouraging parameter and feature sparsity, thereby facilitating robust feature extraction and the assignment of negligible weights to trivial or nonessential features.

Notably, the unit quaternion \( q = (\eta, \zeta) \) is described using a scalar \( \eta \) for the real component and a three-dimensional vector \( \zeta \) for the imaginary component, expressed as:
\begin{equation}
\log \mathbf{q} = \begin{cases}
    \frac{\mathbf{\eta}}{\|\mathbf{\eta}\|} \cos^{-1} \zeta, & \text{if } \|\mathbf{\eta}\| \neq 0 \\
    \mathbf{0}, & \text{otherwise}.
\end{cases}
\end{equation}
The logarithm of quaternion is often utilized in camera pose regression tasks because it supports a smooth and differentiable representation of orientation. Nonetheless, $q$ and $-q$ stand for the same rotation angle because of the dual hemispheres. In this study, we restrict quaternions to a single hemisphere using absolute values to ensure uniqueness.

\begin{table*}[t]
  \centering
   \caption{Camera localization results on Cambridge Landmarks dataset. For each scene, we calculate the average of median errors in both position and rotation of various single-scene and multi-scene camera localization baselines and our proposed method. The most accurate relocalization results are highlighted in \textbf{bold}.
 }
  \footnotesize
  \setlength{\tabcolsep}{0.8pt}
    \begin{tabular}{clccccc}
\toprule
&  & \textbf{King's College}  & \textbf{Old Hospital} & \textbf{Shop Façade} &
\textbf{St Mary's Church}  & \textbf{Average} \\\midrule
\multirow{6}{*}{{\rotatebox[origin=c]{90}{\scriptsize Single Scene}}} 
& PoseNet & 1.94m,5.43\degree & 0.61m,2.92\degree & 1.16m,3.92\degree & 2.67m,8.52\degree & 1.60m,5.20\degree \\
& BayesianPoseNet & 1.76m,4.08\degree & 2.59m,5.18\degree & 1.27m,7.58\degree & 2.13m,8.42\degree & 1.94m,6.32\degree \\
& MapNet & 1.08m,1.91\degree & 1.96m,3.95\degree & 1.51m,4.26\degree & 2.02m,4.57\degree & 1.64m,3.67\degree \\
& PoseNet17 & 1.62m,2.31\degree & 2.64m,3.93\degree & 1.16m,5.77\degree & 2.95m,6.50\degree & 2.09m,4.63\degree \\
& IRPNet & 1.21m,2.19\degree & 1.89m,3.42\degree & 0.74m,3.51\degree & 1.89m,4.98\degree & 1.43m,3.53\degree \\
& PoseNet-Lstm & 0.99m,3.74\degree & 1.53m,4.33\degree & 1.20m,7.48\degree & 1.54m,6.72\degree & 1.32m,5.57\degree \\
\midrule
\multirow{4}{*}{{\rotatebox[origin=c]{90}{\scriptsize Multi-Scene}}} 
& MSPN & 1.77m,3.76\degree & 2.55m,4.05\degree & 2.92m,7.49\degree & 2.67m,6.18\degree & 2.48m,5.37\degree \\
& MS-Trans & 0.85m,\textbf{1.63}\degree & 1.83m,2.43\degree & 0.88m,3.11\degree & 1.64m,4.03\degree & 1.30m,2.80\degree \\
& c2f-MsTrans & 0.71m,2.71\degree & 1.50m,2.98\degree & \textbf{0.61m},\textbf{2.92}\degree & 1.16m,\textbf{3.92}\degree & 0.99m,3.13\degree \\
& MVL-Loc(ours)& \textbf{0.62m},1.89\degree & \textbf{1.38m},\textbf{2.41}\degree & 0.63m,3.22\degree & \textbf{1.09m},4.09\degree & \textbf{0.93m},\textbf{2.90}\degree \\

\bottomrule
\end{tabular}
 \label{tbl:Cambridge}
\end{table*}%

\section{Experiments}
\subsection{Implementation Details}

To ensure consistent network training, input images are cropped to a size of 224 × 224 using random and central cropping techniques.  We utilized a pre-trained CLIP model, which was trained on a diverse set of datasets, including MS-COCO, Visual Genome, and others.
We trained the model from scratch for 280 epochs on an Nvidia V100 GPU using PyTorch 1.11.0. To enhance data diversity, we applied large-scale ColorJitter augmentation by adjusting brightness, contrast, saturation, and hue with respective factors of 0.6, 0.7, 0.7, and 0.5. The model was optimized using the AdamW optimizer, coupled with a cosine learning rate scheduler.
The initial learning rate was set to \( 4.5 \times 10^{-5} \), with a weight decay of \( 4 \times 10^{-5} \). The model was trained using a minibatch size of 64, a dropout rate of 0.5, and weight initializations of \( \alpha = -4.0 \) and \( \beta = -2.0 \).

\subsection{Datasets}

7 Scenes\cite{7scenes} contains RGB-D indoor office environments (3-6 m³) captured by a Kinect camera (640×480 resolution), with camera poses from KinectFusion. Cambridge Landmarks\cite{kendall2015posenet} provides outdoor smartphone-captured images (1920×1080) with SfM-generated poses, covering six scenes under varying lighting, weather, and dynamic obstructions (pedestrians, vehicles).

\subsection{Baselines}

To rigorously validate our framework, we benchmark against both single-scene and multi-scene end-to-end trainable approaches. For the experiments conducted on 7 Scenes dataset, we selected prominent learning-based methods: PoseNet \cite{kendall2015posenet}, BayesianPoseNet \cite{kendall2016modelling}, PoseNet-Lstm \cite{walch2017image}, PoseNet17 \cite{kendall2017geometric}, IRPNet \cite{shavitferensirpnet}, Hourglass \cite{mel2017hourglass}, and AtLoc \cite{wang2020atloc}. Additionally, we validated the model's generalization using the outdoor Cambridge Landmarks dataset, comparing it against several multi-scene state-of-the-art approaches: MSPN \cite{blanton2020extending}, MS-Trans \cite{shavit2021mstrans} and c2f-MsTrans \cite{shavit2023c2f}. For a fair comparative analysis, we selected four benchmark scenes from the Cambridge Landmarks dataset, excluding Great Court and Street. This exclusion was necessary to ensure a balanced comparison of average multi-scene camera relocalization accuracy, as IRPNet, LSTM-PoseNet, and MS-Trans failed to converge on these two scenes.

\subsection{Quantitative Results Analysis}

Table~\ref{tbl:7Scenes} shows MVL-Loc’s superior performance on the 7 Scenes dataset compared to single-scene methods. MVL-Loc achieves a $\mathbf{23.8\%}$ improvement in position accuracy and $\mathbf{10.6\%}$ in rotation accuracy over AtLoc. It reduces position error from $0.22m$ to $0.16m$ and rotation error from $7.32\degree$ to $3.82\degree$ in the $Pumpkin$ scene through precise language descriptions. Compared to the Hourglass model, these descriptions help refine pose estimation and vertical positioning. Our MVL-Loc also outperforms MS-Trans, with $\mathbf{20\%}$ better position accuracy and $\mathbf{5.6\%}$ better rotation accuracy.

On the Cambridge Landmarks dataset, MVL-Loc demonstrates effective generalization in Table \ref{tbl:Cambridge}. It improves position accuracy from $0.71m$ to $0.62m$ on King’s College and from $1.50m$ to $1.38m$ on Old Hospital, and reduces rotation error from $2.71\degree$ to $1.89\degree$ and from $2.98\degree$ to $2.41\degree$ respectively. Language descriptions enhance pose estimation, with notable performance gains of $\mathbf{47.5\%}$ and $\mathbf{29.2\%}$ on Shop Façade and St Mary’s Church compared to PoseNet-LSTM. MVL-Loc effectively combines visual and linguistic cues for robust camera relocalization.

\begin{table*}[t]
  \centering
   \caption{Camera localization results on 7 Scenes dataset. For each scene, we calculate the average of median errors in both position and rotation of various single-scene and multi-scene camera localization baselines and our proposed method. The most accurate relocalization results are highlighted in \textbf{bold}.}
  \scriptsize
  \setlength{\tabcolsep}{0.6pt}
    \begin{tabular}{clcccccccc}
\toprule
&  & \textbf{Chess} & \textbf{Fire} & \textbf{Heads} & \textbf{Office} &
\textbf{Pumpkin} & \textbf{Kitchen} & \textbf{Stairs} & \textbf{Average} \\\midrule
\multirow{7}{*}{{\rotatebox[origin=c]{90}{\scriptsize Single-Scene}}} & PoseNet 
& 0.32m,7.60\degree & 0.48m,14.6\degree & 0.31m,12.2\degree
& 0.48m,7.68\degree & 0.47m,8.42\degree & 0.59m,8.64\degree & 0.47m,13.81\degree
& 0.45m,10.42\degree\\
& Bayesian & 0.38m,7.24\degree & 0.43m,13.8\degree
& 0.30m,12.3\degree & 0.49m,8.09\degree & 0.63m,7.18\degree & 0.59m,7.59\degree
& 0.48m,13.22\degree & 0.47m,9.91\degree \\
& PN-Lstm & 0.24m,5.79\degree & 0.34m,12.0\degree &
0.22m,13.8\degree & 0.31m,8.11\degree & 0.34m,7.03\degree & 0.37m,8.83\degree &
0.41m,13.21\degree & 0.32m,9.82\degree \\
& PoseNet17 & 0.14m,4.53\degree & 0.29m,11.5\degree &
0.19m,13.1\degree & 0.20m,5.62\degree & 0.27m,4.77\degree & 0.24m,5.37\degree &
0.36m,12.53\degree & 0.24m,8.20\degree \\
& IRPNet & 0.13m,5.78\degree & 0.27m,9.83\degree &
0.17m,13.2\degree & 0.25m,6.41\degree & 0.23m,5.83\degree & 0.31m,7.32\degree &
0.35m,11.91\degree & 0.24m,8.61\degree\\
& Hourglass & 0.15m,6.18\degree & 0.27m,10.83\degree
& 0.20m,\textbf{11.6}\degree & 0.26m,8.59\degree & 0.26m,7.32\degree & 0.29m,10.7\degree 
& 0.30m,12.75\degree & 0.25m,9.71\degree \\
& AtLoc & 0.11m,4.37\degree & 0.27m,11.7\degree
& 0.16m,11.9\degree & 0.19m,\textbf{5.61}\degree & 0.22m,4.54\degree 
& 0.25m,\textbf{5.62}\degree & 0.28m,10.9\degree & 0.21m,7.81\degree \\
\midrule
\multirow{4}{*}{{\rotatebox[origin=c]{90}{\scriptsize MultiScene}}} 
& MSPN & 0.10m,4.76\degree & 0.29m,11.5\degree &
0.17m,13.2\degree & 0.17m,6.87\degree & 0.21m,5.53\degree & 0.23m,6.81\degree & 0.31m,11.81\degree & 0.21m,8.64\degree \\
& MS-Trans & 0.11m,4.67\degree & 0.26m,9.78\degree 
& 0.16m,12.8\degree & 0.17m,5.66\degree & 0.18m,4.44\degree & 0.21m,5.99\degree 
& 0.29m,8.45\degree & 0.20m,7.40\degree \\
& c2f-MsTrans & 0.10m,4.63\degree & 0.25m,9.89\degree & 0.14m,12.5\degree &
0.16m,5.65\degree & \textbf{0.16m},4.42\degree & 0.18m,6.29\degree & 0.27m,\textbf{7.86\degree} & 0.18m,7.32\degree \\
& \textbf{MVL-Loc} & \textbf{0.09m},\textbf{3.95\degree} & \textbf{0.22m},\textbf{9.45}\degree & \textbf{0.11m},11.9\degree &
\textbf{0.14m},5.68\degree & \textbf{0.16m},\textbf{3.82}\degree & \textbf{0.14m},6.11\degree & \textbf{0.23m},8.11\degree &
\textbf{0.16m},\textbf{6.98}\degree\\
\bottomrule
\end{tabular}
 \label{tbl:7Scenes}
\end{table*}

\begin{figure}[tbh]
  \centering
    \caption{Attention visualizations on \textit{Heads} and \textit{Fire} scenes with 1, 3, and 7-scene training.}
  \begin{subfigure}[b]{0.155\textwidth}
    \centering
    \captionsetup{justification=centering}
    \includegraphics[width=\linewidth]{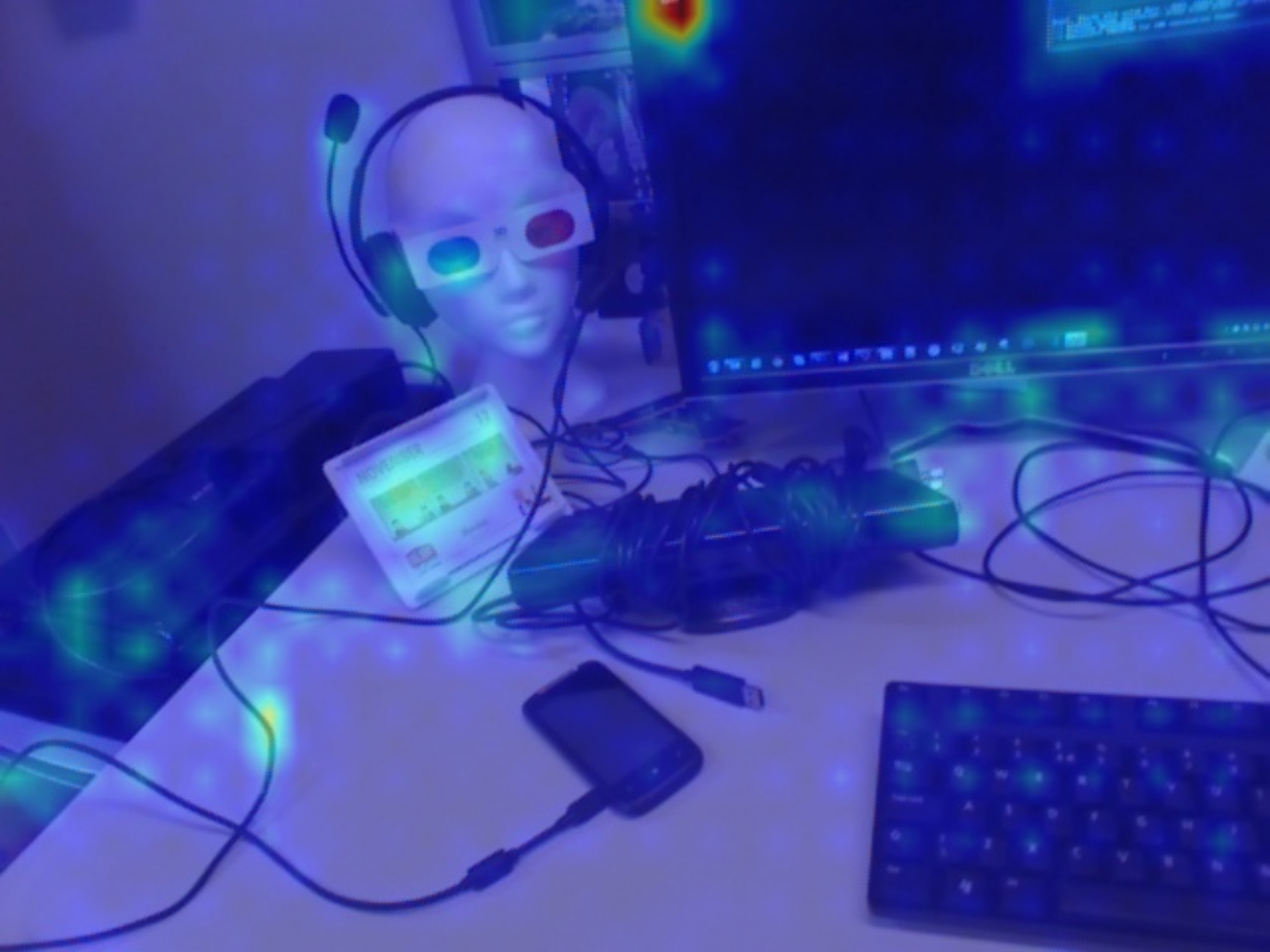}
    \caption{Heads\\(1 scene)}
    \label{fig:headsubfig1}
  \end{subfigure}%
  \begin{subfigure}[b]{0.155\textwidth}
    \centering
    \captionsetup{justification=centering}
    \includegraphics[width=\linewidth]{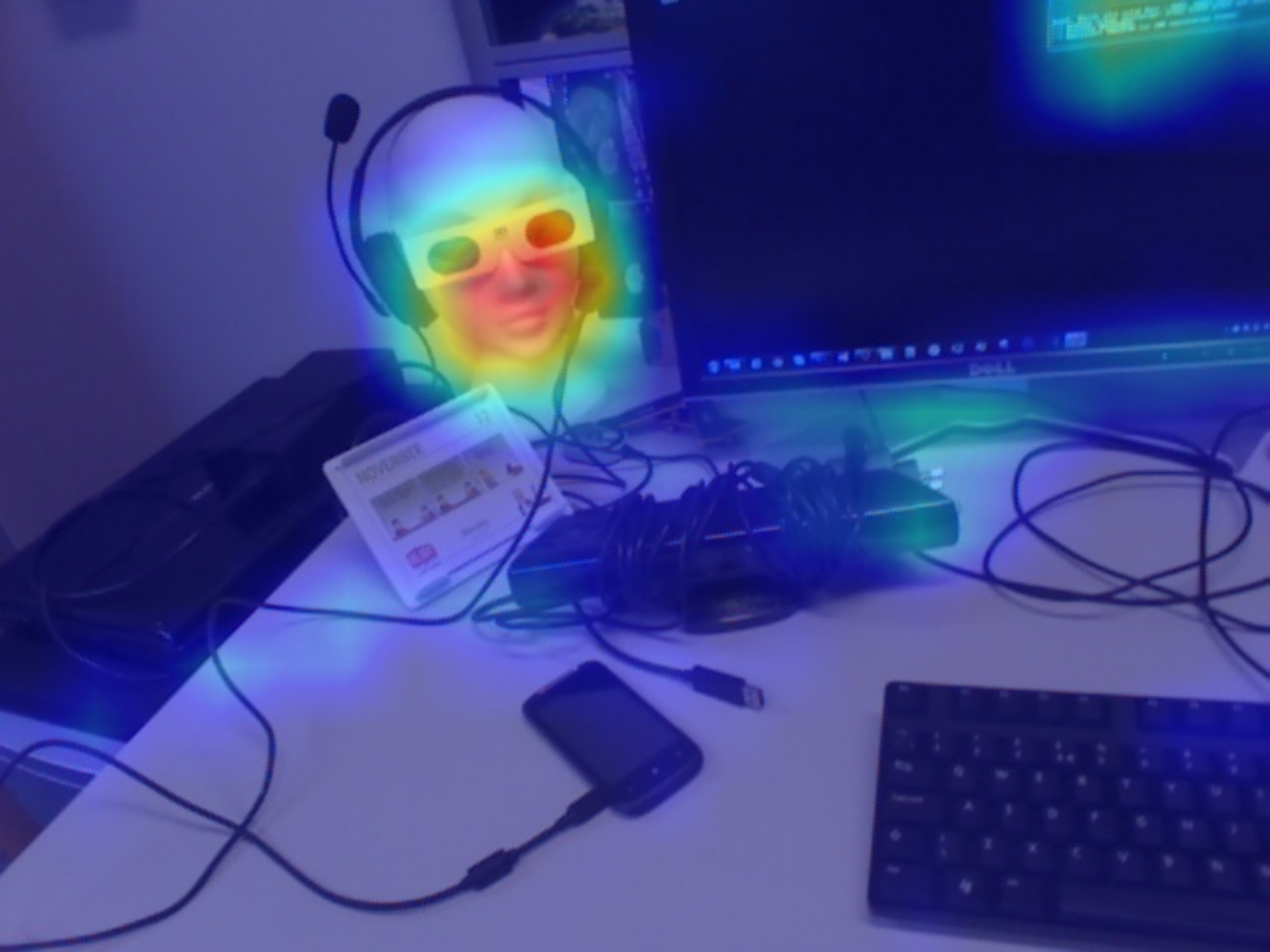}
    \caption{Heads\\(3 scenes)}
    \label{fig:headsubfig2}
  \end{subfigure}%
  \begin{subfigure}[b]{0.155\textwidth}
    \centering
    \captionsetup{justification=centering}
    \includegraphics[width=\linewidth]{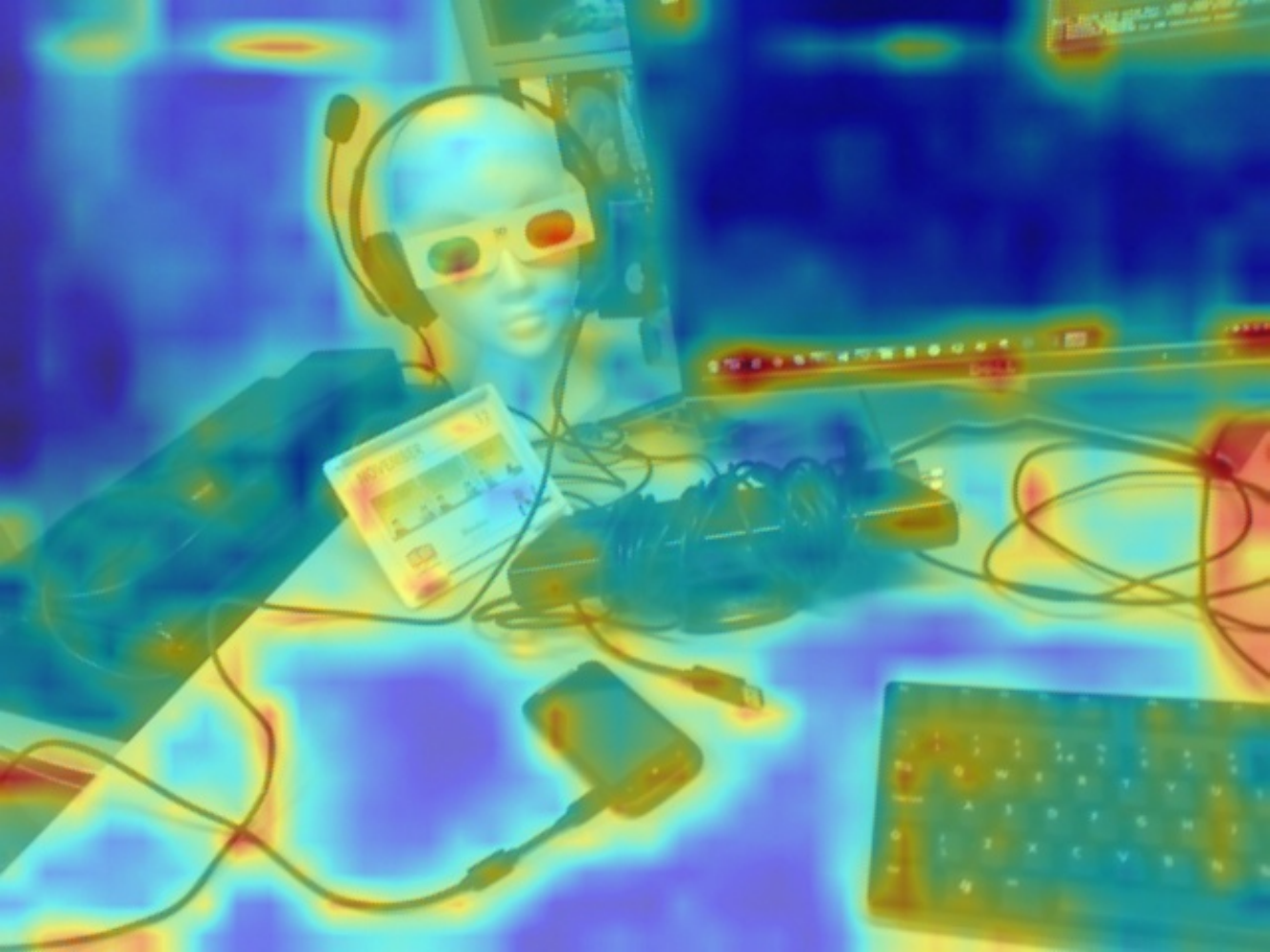}
    \caption{Heads\\(7 scenes)}
    \label{fig:headsubfig3}
  \end{subfigure}%
  \begin{subfigure}[b]{0.155\textwidth}
    \centering
    \captionsetup{justification=centering}
    \includegraphics[width=\linewidth]{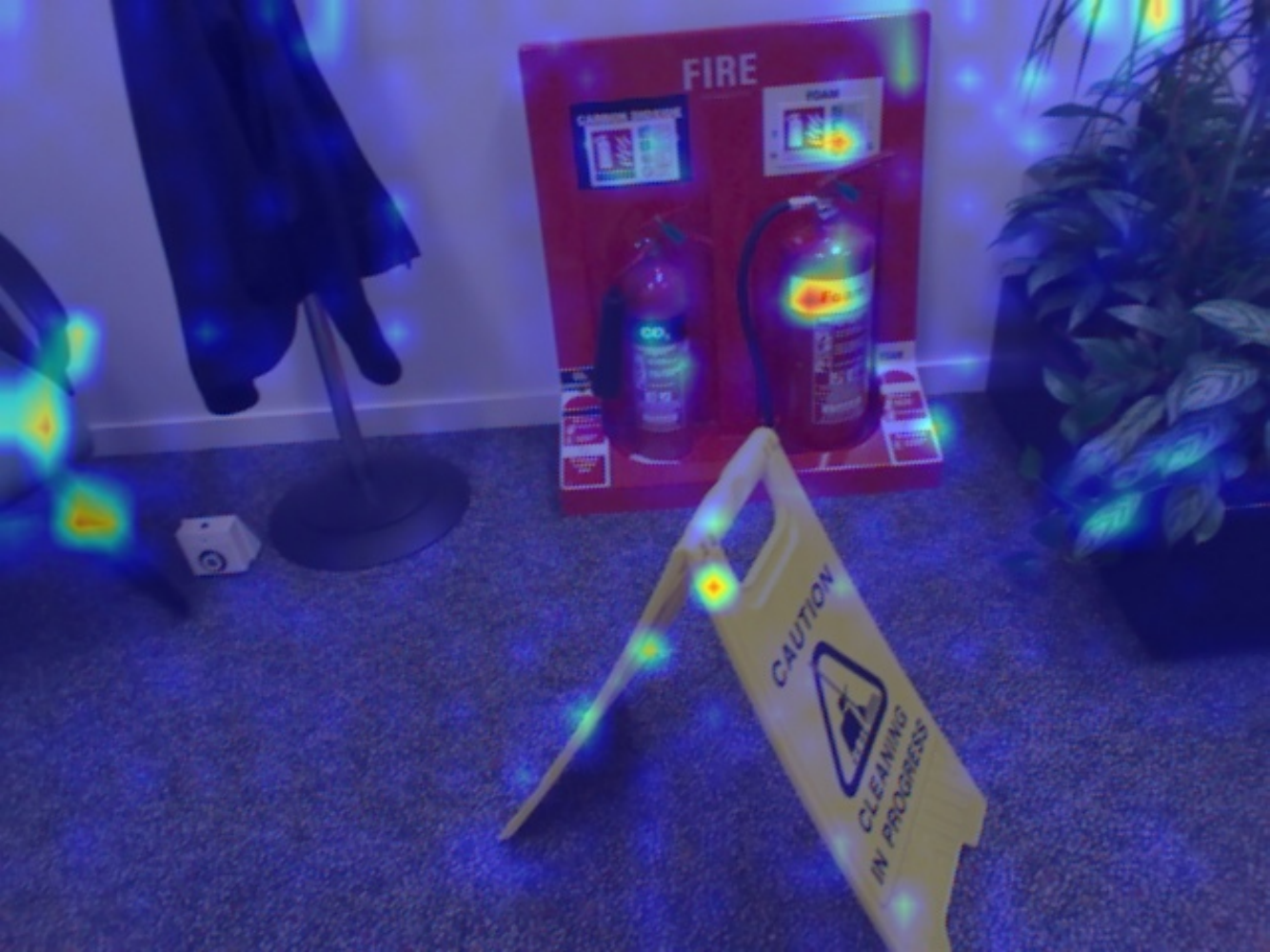}
    \caption{Fire\\(1 scene)}
    \label{fig:firesubfig1}
  \end{subfigure}%
  \begin{subfigure}[b]{0.155\textwidth}
    \centering
    \captionsetup{justification=centering}
    \includegraphics[width=\linewidth]{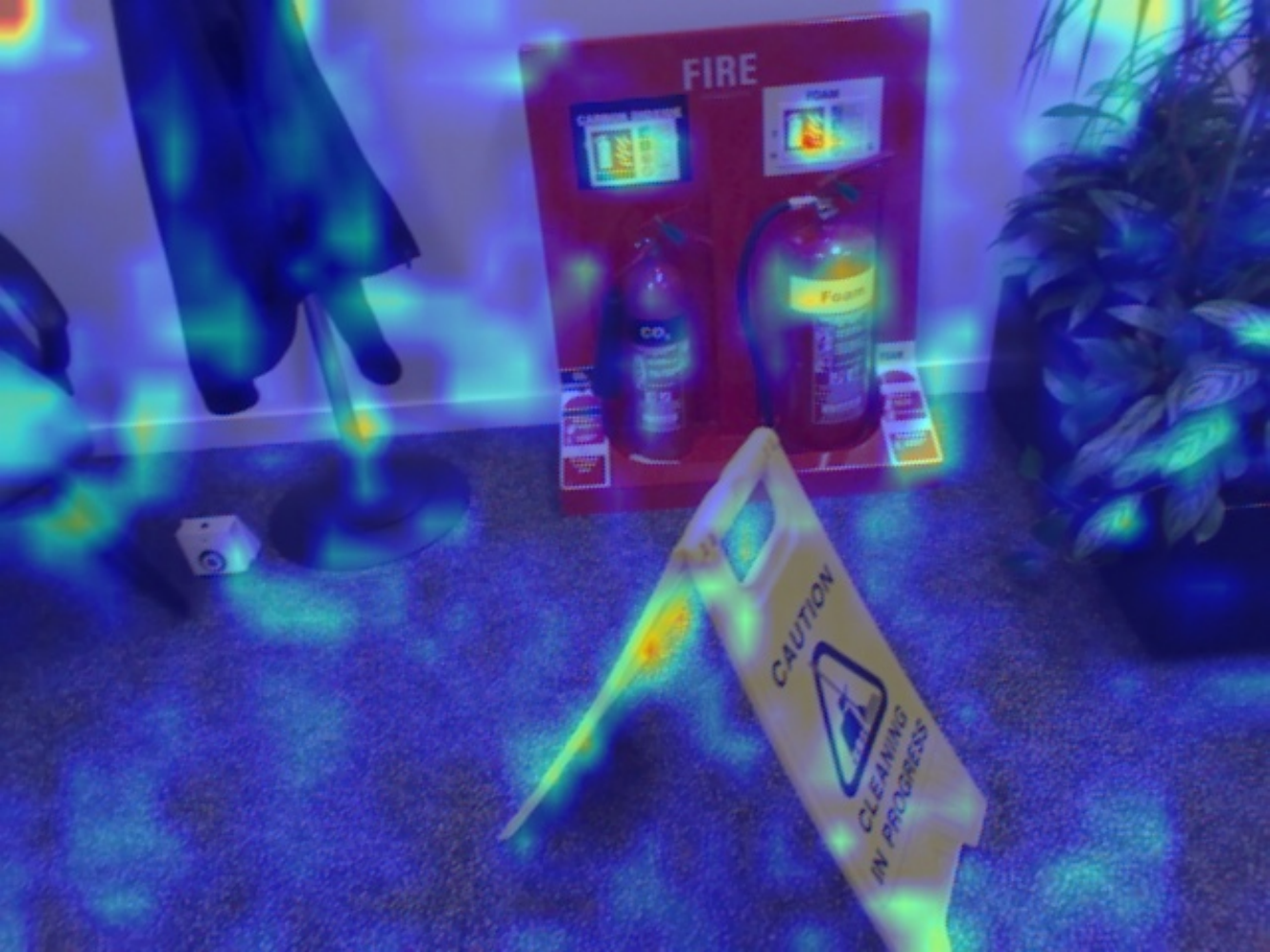}
    \caption{Fire\\(3 scenes)}
    \label{fig:firesubfig2}
  \end{subfigure}%
  \begin{subfigure}[b]{0.155\textwidth}
    \centering
    \captionsetup{justification=centering}
    \includegraphics[width=\linewidth]{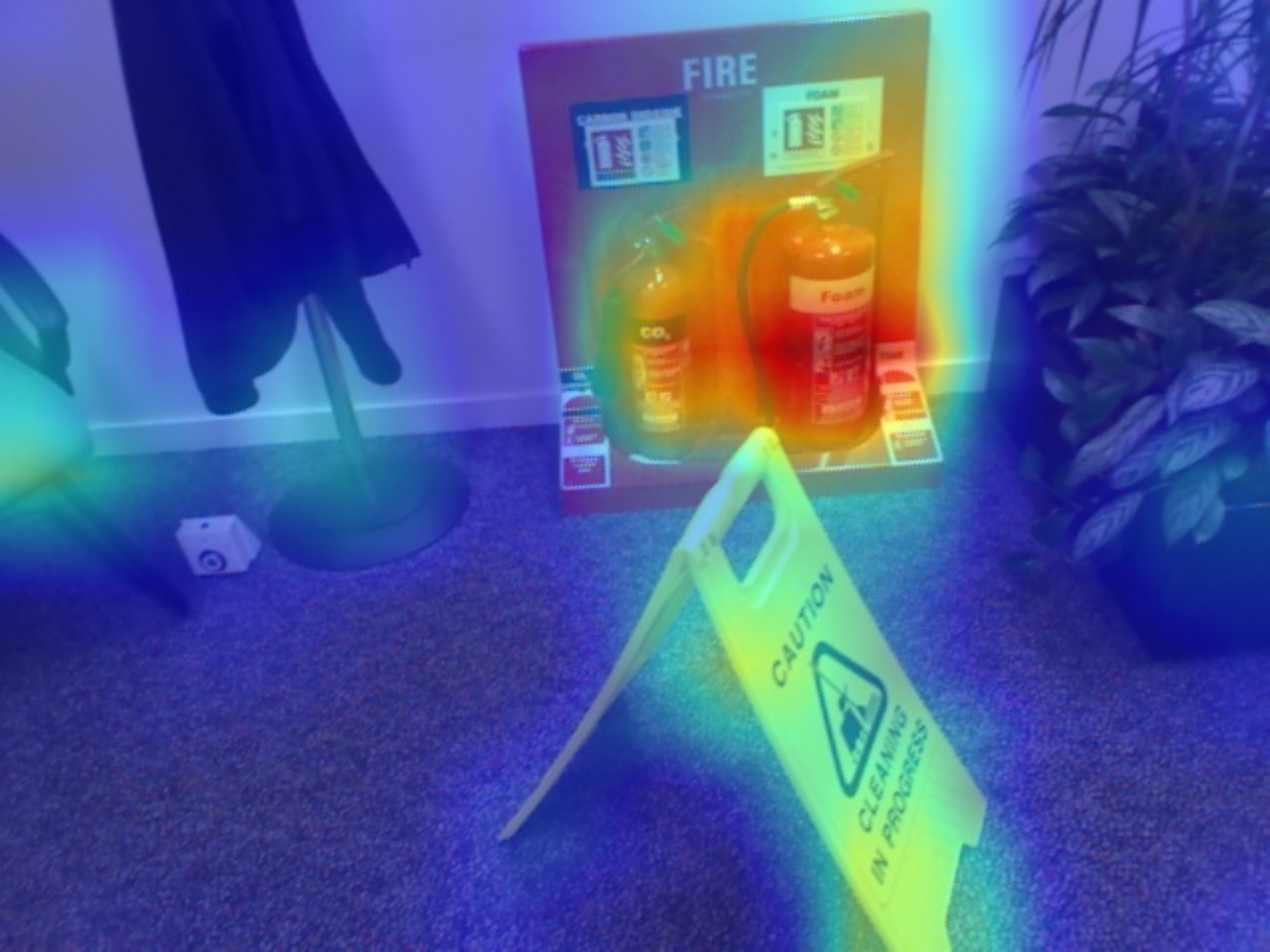}
    \caption{Fire\\(7 scenes)}
    \label{fig:firesubfig3}
  \end{subfigure}
  \label{fig:multi_scene_attention}
\end{figure}

\begin{figure}[t]
  \centering
\caption{Saliency maps on \textit{King's College} and \textit{St Mary's Church}. From left to right: original images, MS-Trans, and MVL-Loc (ours). MVL-Loc shows sharper focus on  key architectural structures, aiding more accurate pose estimation.}
  \begin{subfigure}[b]{0.3\columnwidth}
    \centering
    \includegraphics[width=\linewidth,height=0.7\columnwidth]{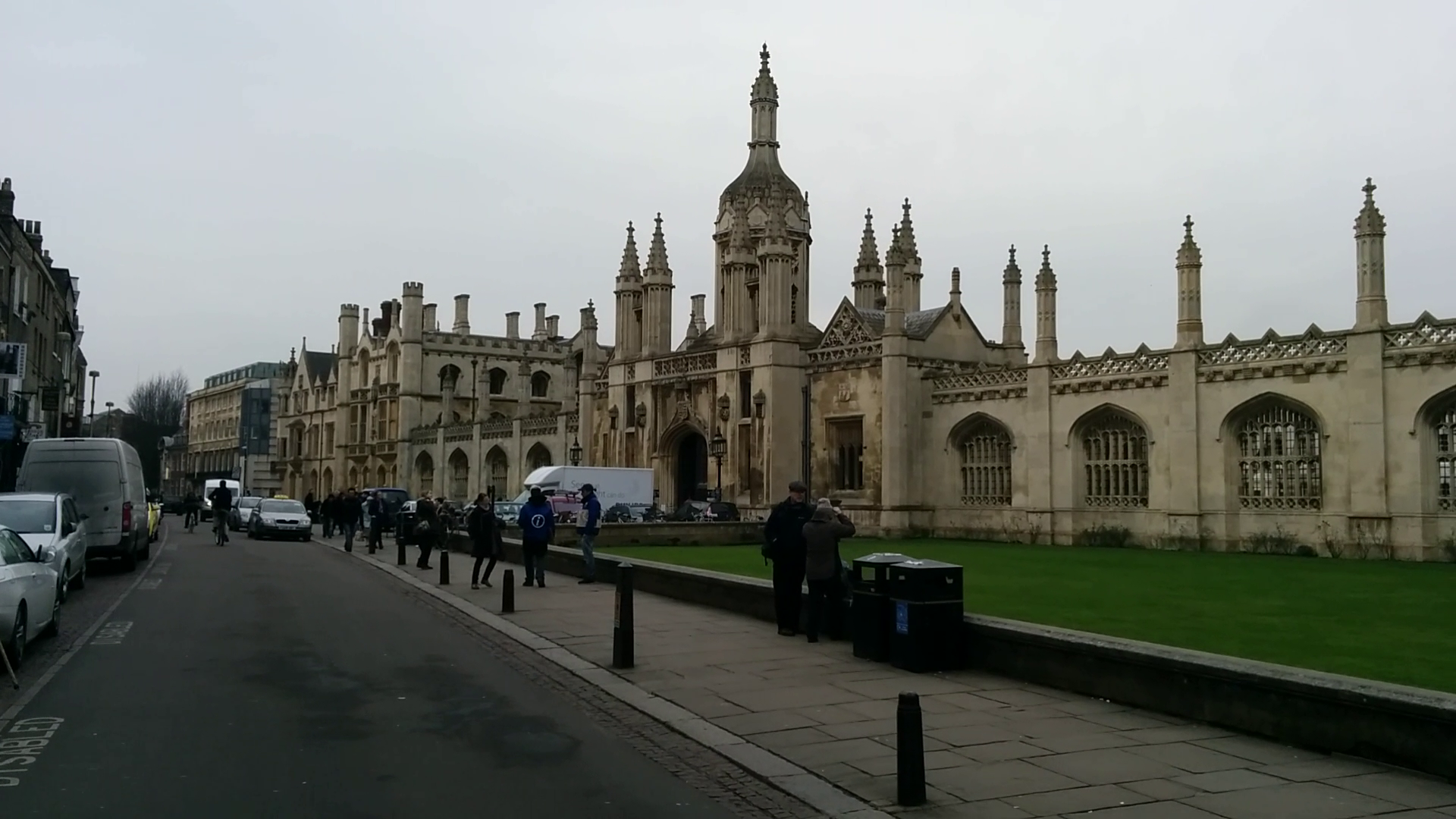}
    \caption*{Original}
  \end{subfigure}%
  \hspace{0.01\columnwidth}
  \begin{subfigure}[b]{0.3\columnwidth}
    \centering
    \includegraphics[width=\linewidth,height=0.7\columnwidth]{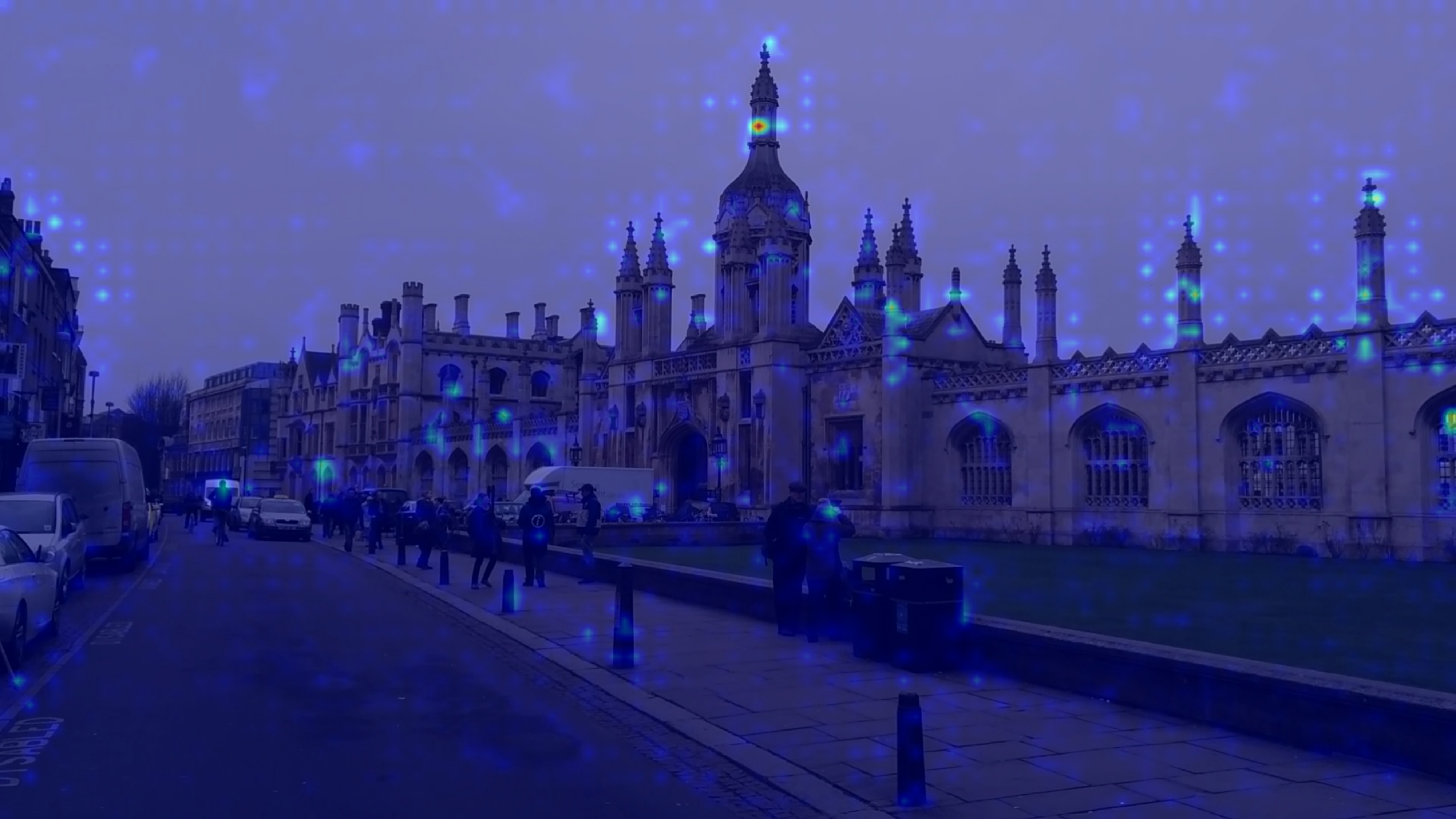}
    \caption*{MS-Trans}
  \end{subfigure}%
  \hspace{0.01\columnwidth}
  \begin{subfigure}[b]{0.3\columnwidth}
    \centering
    \includegraphics[width=\linewidth,height=0.7\columnwidth]{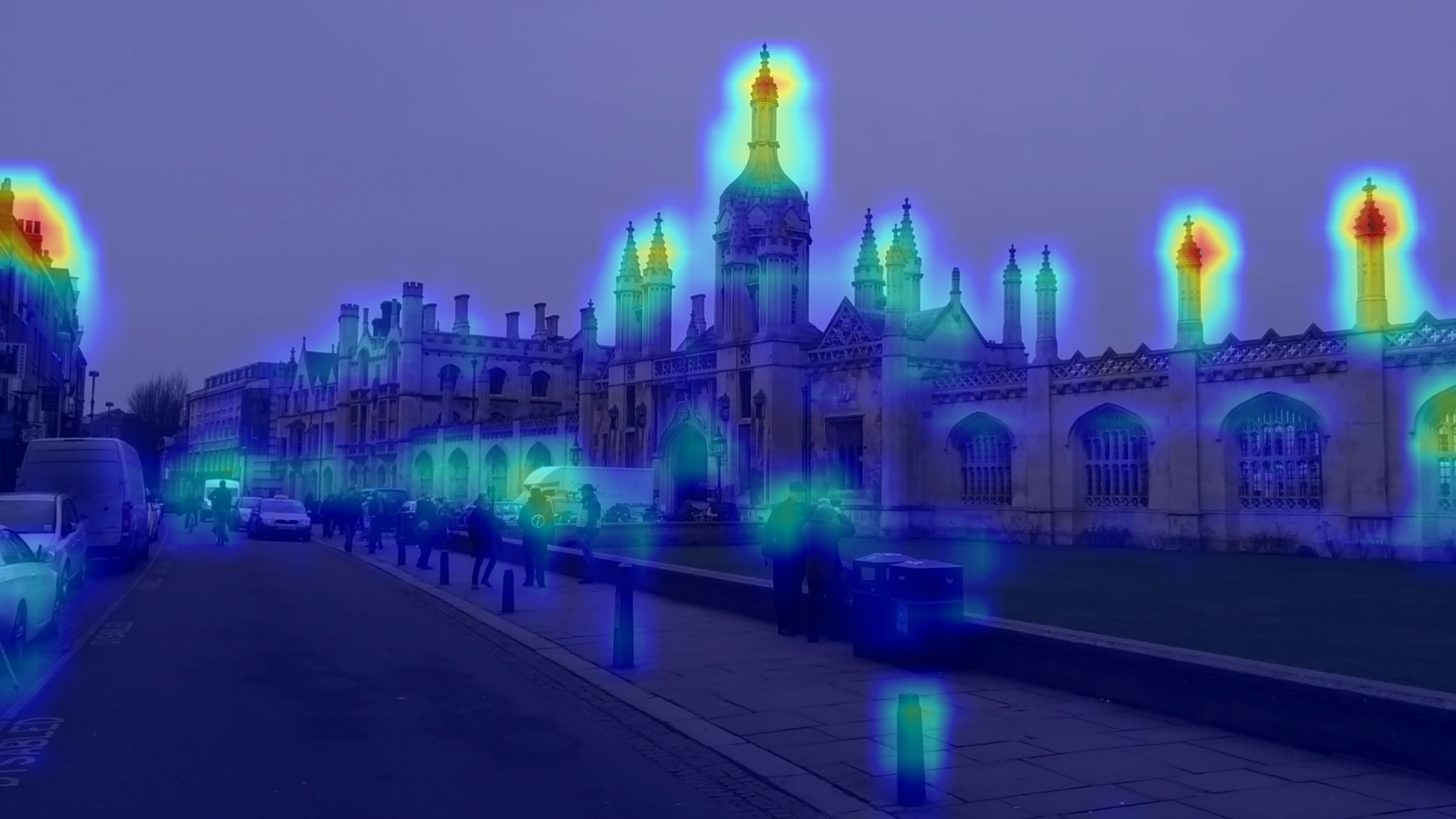}
    \caption*{MVL-Loc (Ours)}
  \end{subfigure}
  \vspace{0.5em}
  \begin{subfigure}[b]{0.3\columnwidth}
    \centering
    \includegraphics[width=\linewidth,height=0.7\columnwidth]{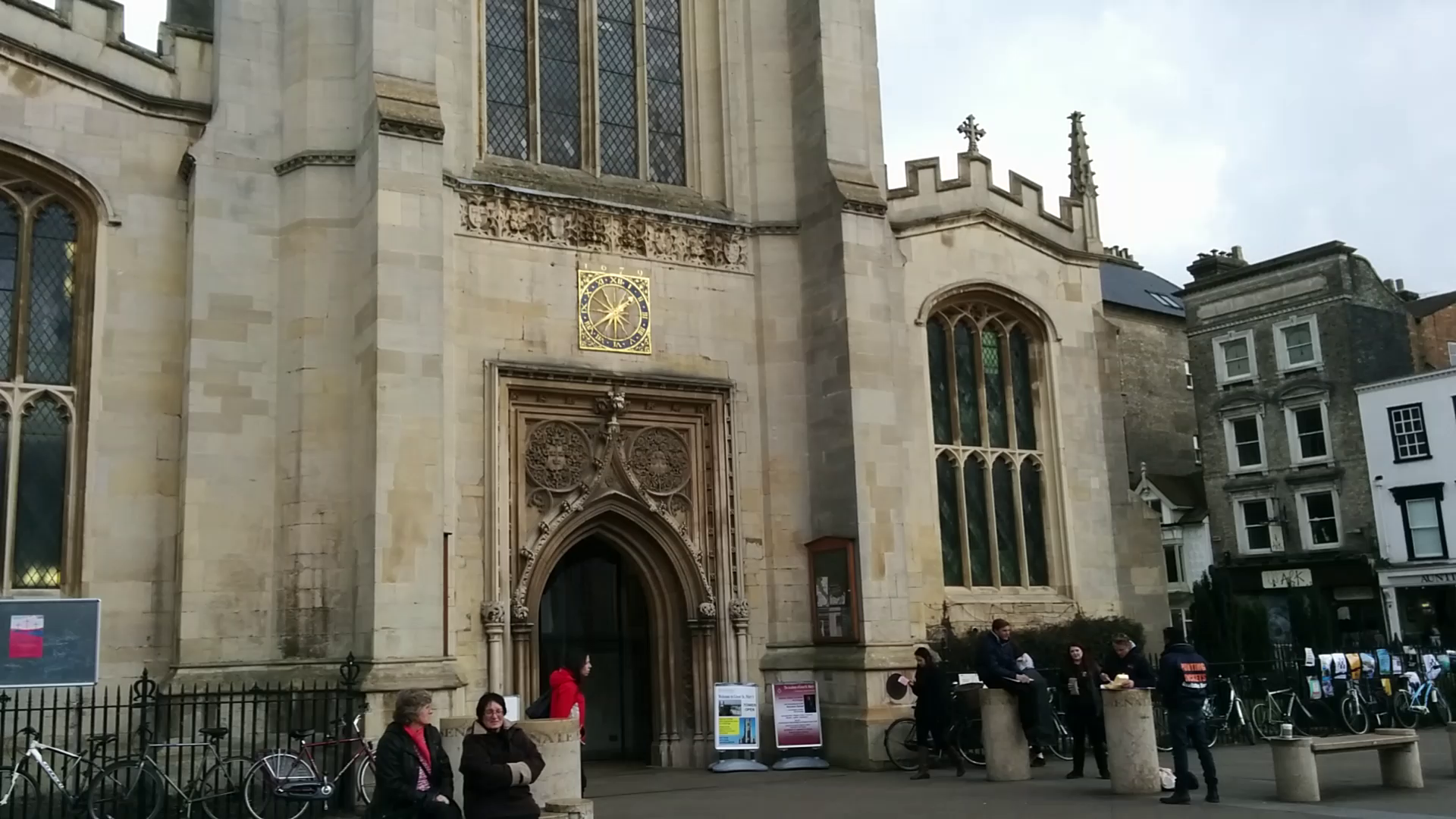}
    \caption*{Original}
  \end{subfigure}%
  \hspace{0.01\columnwidth}
  \begin{subfigure}[b]{0.3\columnwidth}
    \centering
    \includegraphics[width=\linewidth,height=0.7\columnwidth]{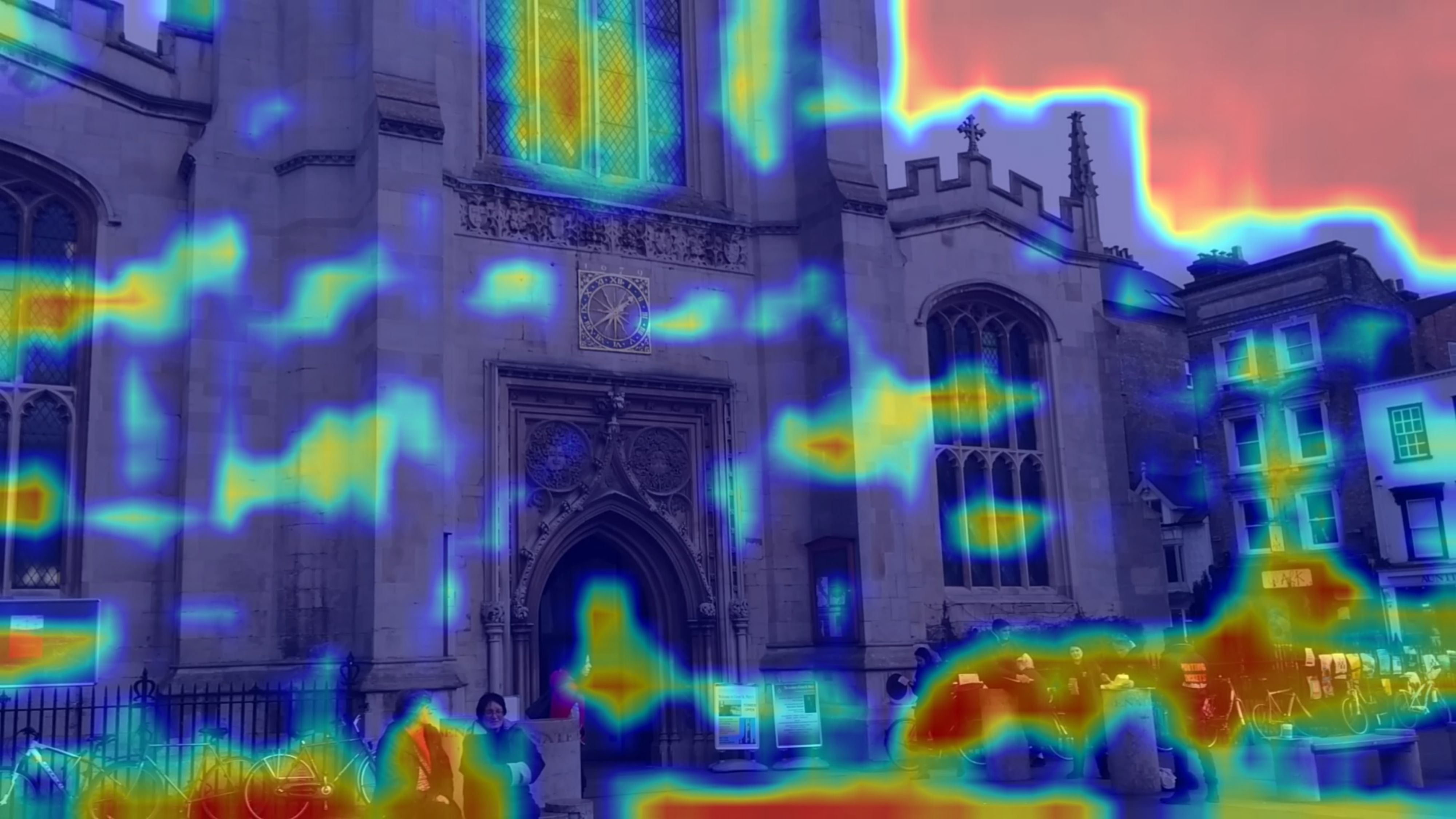}
    \caption*{MS-Trans}
  \end{subfigure}%
  \hspace{0.01\columnwidth}
  \begin{subfigure}[b]{0.3\columnwidth}
    \centering
    \includegraphics[width=\linewidth,height=0.7\columnwidth]{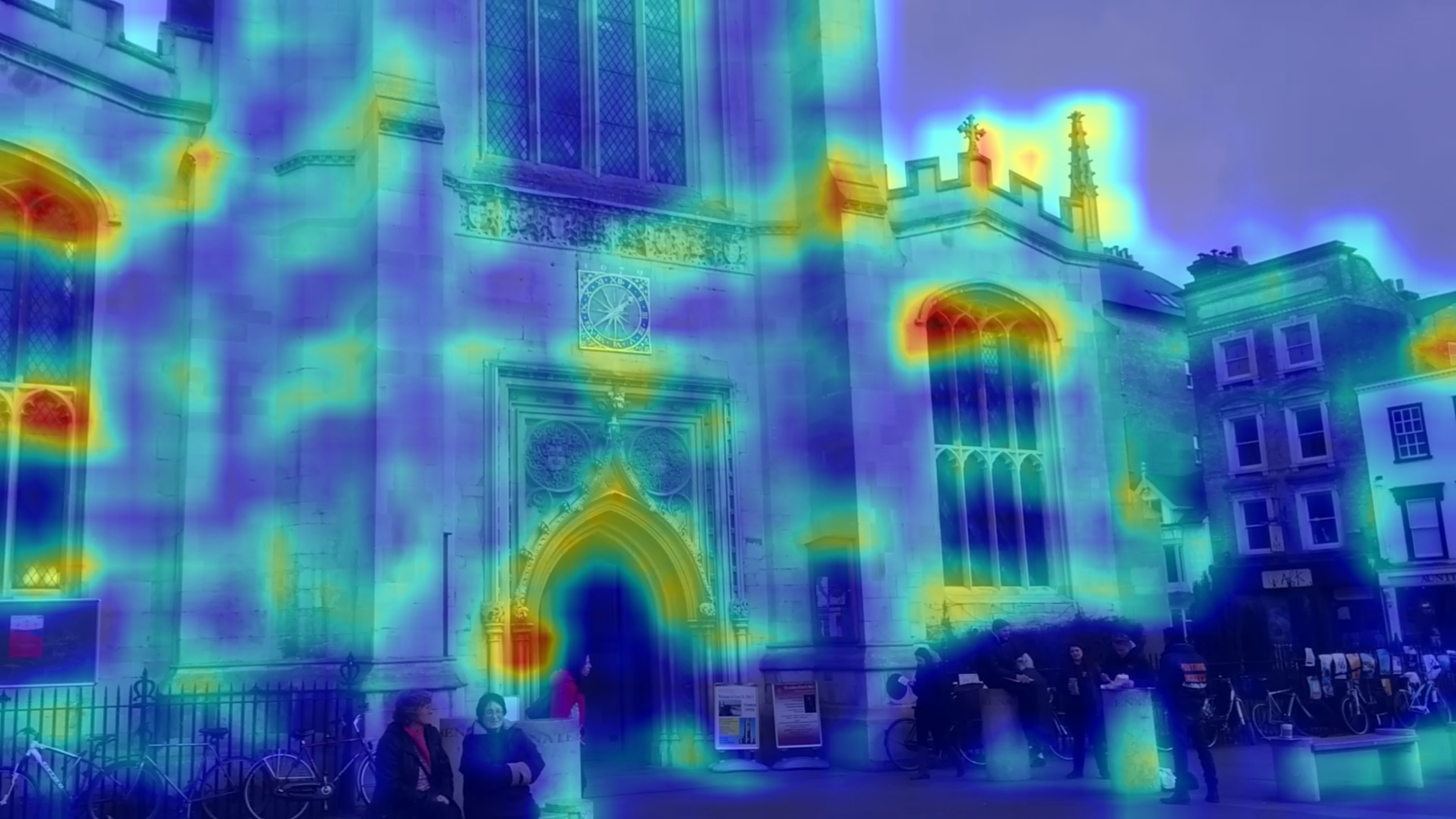}
    \caption*{MVL-Loc (Ours)}
  \end{subfigure}
\label{fig:saliency}
\end{figure}

\begin{table*}[t]
\caption{Ablation study evaluating the effects of World Knowledge (WK), Language Descriptions (LD), and Multi-scene Training (MT) on camera relocalization performance across the 7Scenes and Cambridge Landmarks datasets. }
\label{tb:worldknowledge}%
\centering
\begin{tabular}
{cccccc}
\toprule
\textbf{Pre-trained} & \textbf{WK} & \textbf{LD} & \textbf{MT} & \textbf{7Scenes} &  \textbf{Cambridge Landmarks} \\ \midrule
ImageNet & & & & 0.28m,7.97\degree & 1.63m,4.76\degree \\
Clip & \checkmark &  & & 0.22m,7.38\degree & 1.33m,3.57\degree \\
Clip & \checkmark & \checkmark &  & 0.18m,7.10\degree & 1.01m,3.05\degree \\
Clip & \checkmark& \checkmark & \checkmark & 0.16m,6.98\degree & 0.93m,2.90\degree\\
\bottomrule
\end{tabular}
\end{table*}

\begin{table}[tbh]
\caption{Ablation study evaluating the performance of different vision-language encoders in our model, tested on the 7Scenes and Cambridge Landmarks datasets. }
\label{table:encoder}\centering%
\begin{tabular}{lcc}
\toprule
\textbf{Vision-Language Encoder} & \textbf{7Scenes} & \textbf{Cambridge Land.} \\
\midrule
{BLIP-2} & 0.18m,7.13\degree & 0.98m,3.16\degree \\
{OpenFlamingo} & 0.19m,7.21\degree & 1.03m,3.19\degree \\
{\textbf{Clip}} & 0.16m,6.98\degree & 0.93m,2.90\degree \\\bottomrule
\end{tabular}%
\end{table}

\begin{table}[tbh]
\caption{Ablation study on the impact of varying the number of decoder layers in the Transformer architecture, evaluated using the 7Scenes and Cambridge Landmarks datasets. The table reports the average median position and orientation errors across all scenes. The model with 4 decoder layers, highlighted in bold, represents the chosen configuration due to its balanced performance in both datasets.}
\label{table:decoder}\centering%
\begin{tabular}{ccc}
\toprule
\textbf{Decoder Layers} & \textbf{7Scenes} & \textbf{Cambridge Landmarks} \\
\midrule
{2} & {0.54m,8.15\degree} & {1.71m,3.93\degree} \\
{\textbf{4}} & {0.16m,6.98\degree} & {0.93m,2.90\degree} \\
{6} & {0.17m,7.02\degree} & {0.97m,2.99\degree} \\
{8} & {0.17m,7.16\degree} & {0.98m,3.01\degree} \\ \bottomrule
\end{tabular}%
\end{table}

\subsection{Visualization of Multi-Scene Fusion}

Training on single scenes limits model generalization and scene adaptability. Multi-scene training, however, enables MVL-Loc to recognize key objects, interpret spatial relationships, and generalize across diverse environments. Figure~\ref{fig:multi_scene_attention} visualizes attention in the \textit{Heads} and \textit{Fire} scenes from the 7Scenes dataset, showing activations for one, three, and seven scenes (left to right columns). In \textit{Heads}, single-scene training struggles to extract positional information, with attention sparsely distributed. With three scenes, the model starts identifying stable objects like the face and headset, while seven-scene fusion captures additional elements like the monitor and mannequin head, enhancing spatial understanding. For the \textit{Fire} scene, single-scene training focuses on isolated regions, such as fire extinguishers, while three-scene input expands attention to the umbrella and background elements. Seven-scene fusion integrates local and global cues, improving object contextualization and scene comprehension.

On the Cambridge Landmarks dataset, MVL-Loc, guided by language prompts, outperforms MS-Trans by focusing on static, geometrically significant objects (e.g., spires and arches) in the \textit{College} and \textit{Church} scenes shown in  Figure~\ref{fig:saliency}. MS-Trans, by contrast, focuses on dynamic objects like pedestrians and cars, which are less reliable for relocalization. MVL-Loc’s attention on structural elements like archways and carvings leads to better spatial awareness and enhanced relocalization across diverse environments.

\section{Ablation Study}

We evaluate model components for camera relocalization on 7Scenes and Cambridge Landmarks. Table \ref{tb:worldknowledge} shows the effect of incorporating world knowledge, language descriptions, and multi-scene training. Initially, using ImageNet as a pre-trained model led to poor generalization and weaker relocalization. Adding CLIP improved accuracy to 1.33m and 3.57° in Cambridge Landmarks. Integrating language descriptions further enhanced performance, achieving 0.18m, 7.10° in 7Scenes and 1.01m, 3.05° in Cambridge Landmarks. Multi-scene integration of all components achieves optimal accuracy underscores world knowledge, language guidance, and diverse scene exposure.

Table \ref{table:encoder} compares vision-language encoders: BLIP-2, OpenFlamingo, and CLIP. BLIP-2 underperformed due to limited geometric cue capture. OpenFlamingo showed improvements with average errors of 0.19m, 7.21° in 7Scenes and 1.03m, 3.19° in Cambridge Landmarks. CLIP achieved the highest accuracy—0.16m, 6.98° in 7Scenes and 0.93m, 2.90° in Cambridge Landmarks—demonstrating superior semantic and geometric understanding.

We also tested the impact of decoder layers on performance in Table \ref{table:decoder}. With 2 layers, the model had high errors (0.54m, 8.15° in 7Scenes and 1.71m, 3.93° in Cambridge Landmarks). The optimal performance was achieved with 4 decoder layers, balancing complexity and feature capture effectively.

\section{CONCLUSIONS}

In this work, we propose MVL-Loc, a novel 6-DoF camera relocalization framework that leverages pretrained world knowledge from vision-language models (VLMs) and integrates multimodal data. The architecture generalizes effectively across both indoor and outdoor environments, demonstrating robust performance in various real-world settings. We employ natural language as a directive tool to guide multi-scene learning, allowing the model to capture both semantic information and spatial relationships among objects. Through extensive experiments, MVL-Loc achieves state-of-the-art performance in end-to-end multi-scene camera relocalization.
Looking forward, we plan to explore integrating large language models, such as GPT-o1, with MVL-Loc to enable autonomous scene comprehension. This advancement has the potential to further enhance the precision and adaptability of camera relocalization in complex, real-world environments.
\bibliographystyle{IEEEtran}
\bibliography{root}

\begin{thebibliography}{10}
\providecommand{\url}[1]{#1}
\csname url@samestyle\endcsname
\providecommand{\newblock}{\relax}
\providecommand{\bibinfo}[2]{#2}
\providecommand{\BIBentrySTDinterwordspacing}{\spaceskip=0pt\relax}
\providecommand{\BIBentryALTinterwordstretchfactor}{4}
\providecommand{\BIBentryALTinterwordspacing}{\spaceskip=\fontdimen2\font plus
\BIBentryALTinterwordstretchfactor\fontdimen3\font minus \fontdimen4\font\relax}
\providecommand{\BIBforeignlanguage}[2]{{%
\expandafter\ifx\csname l@#1\endcsname\relax
\typeout{** WARNING: IEEEtran.bst: No hyphenation pattern has been}%
\typeout{** loaded for the language `#1'. Using the pattern for}%
\typeout{** the default language instead.}%
\else
\language=\csname l@#1\endcsname
\fi
#2}}
\providecommand{\BIBdecl}{\relax}
\BIBdecl

\bibitem{fischler1981random}
M.~A. Fischler and R.~C. Bolles, ``Random sample consensus: A paradigm for model fitting with applications to image analysis and automated cartography,'' \emph{Commun. ACM}, vol.~24, no.~6, p. 381–395, Jun. 1981.

\bibitem{kendall2015posenet}
A.~Kendall, M.~Grimes, and R.~Cipolla, ``Posenet: A convolutional network for real-time 6-dof camera relocalization,'' in \emph{CVPR}, 2015.

\bibitem{walch2017image}
F.~Walch, C.~Hazirbas, L.~Leal-Taixe, T.~Sattler, S.~Hilsenbeck, and D.~Cremers, ``Image-based localization using lstms for structured feature correlation,'' in \emph{CVPR}, 2017.

\bibitem{melekhov2017image}
I.~Melekhov, J.~Ylioinas, J.~Kannala, and E.~Rahtu, ``Image-based localization using hourglass networks,'' in \emph{ICCV}, 2017.

\bibitem{kendall2016modelling}
A.~Kendall and R.~Cipolla, ``Modelling uncertainty in deep learning for camera relocalization,'' in \emph{ICRA}, 2016.

\bibitem{kendall2017geometric}
------, ``Geometric loss functions for camera pose regression with deep learning,'' in \emph{CVPR}, 2017.

\bibitem{clark2017vidloc}
R.~Clark, S.~Wang, A.~Markham, N.~Trigoni, and H.~Wen, ``Vidloc: A deep spatio-temporal model for 6-dof video-clip relocalization,'' in \emph{CVPR}, 2017.

\bibitem{wang2020atloc}
B.~Wang, C.~Chen, C.~X. Lu, P.~Zhao, N.~Trigoni, and A.~Markham, ``Atloc: Attention guided camera localization,'' in \emph{Proceedings of the AAAI Conference on Artificial Intelligence}, vol.~34, no.~06, 2020, pp. 10\,393--10\,401.

\bibitem{xiao2024effloc}
Z.~Xiao, C.~Chen, S.~Yang, and W.~Wei, ``Effloc: Lightweight vision transformer for efficient 6-dof camera relocalization,'' in \emph{2024 IEEE International Conference on Robotics and Automation (ICRA)}, 2024, pp. 8529--8536.

\bibitem{zhou2023navgpt}
G.~Zhou, Y.~Hong, and Q.~Wu, ``Navgpt: Explicit reasoning in vision-and-language navigation with large language models,'' \emph{arXiv preprint arXiv:2305.16986}, 2023.

\bibitem{rt22023arxiv}
A.~Brohan, N.~Brown, J.~Carbajal, Y.~Chebotar, and X.~Chen, ``Rt-2: Vision-language-action models transfer web knowledge to robotic control,'' in \emph{arXiv preprint arXiv:2307.15818}, 2023.

\bibitem{blanton2020extending}
H.~Blanton, C.~Greenwell, S.~Workman, and N.~Jacobs, ``Extending absolute pose regression to multiple scenes,'' in \emph{Proceedings of the IEEE/CVF Conference on Computer Vision and Pattern Recognition Workshops}, 2020, pp. 38--39.

\bibitem{shavit2023c2f}
Y.~Shavit, R.~Ferens, and Y.~Keller, ``Coarse-to-fine multi-scene pose regression with transformers,'' \emph{IEEE transactions on pattern analysis and machine intelligence}, vol.~PP, 08 2023.

\bibitem{chen2011city}
D.~M. Chen, G.~Baatz, K.~K{\"o}ser, S.~S. Tsai, R.~Vedantham, T.~Pylv{\"a}n{\"a}inen, K.~Roimela, X.~Chen, J.~Bach, M.~Pollefeys \emph{et~al.}, ``City-scale landmark identification on mobile devices,'' in \emph{CVPR}, 2011.

\bibitem{shen2025imagdressing}
F.~Shen, X.~Jiang, X.~He, H.~Ye, C.~Wang, X.~Du, Z.~Li, and J.~Tang, ``Imagdressing-v1: Customizable virtual dressing,'' in \emph{Proceedings of the AAAI Conference on Artificial Intelligence}, vol.~39, no.~7, 2025, pp. 6795--6804.

\bibitem{shen2024imagpose}
F.~Shen and J.~Tang, ``Imagpose: A unified conditional framework for pose-guided person generation,'' \emph{Advances in neural information processing systems}, vol.~37, pp. 6246--6266, 2024.

\bibitem{brachmann2017dsac}
E.~Brachmann, A.~Krull, S.~Nowozin, J.~Shotton, F.~Michel, S.~Gumhold, and C.~Rother, ``Dsac-differentiable ransac for camera localization,'' in \emph{CVPR}, 2017.

\bibitem{camposeco2019hybrid}
F.~Camposeco, A.~Cohen, M.~Pollefeys, and T.~Sattler, ``Hybrid scene compression for visual localization,'' in \emph{CVPR}, 2019.

\bibitem{arnold2022mapfree}
E.~Arnold, J.~Wynn, S.~Vicente, G.~Garcia-Hernando, {\'{A}}.~Monszpart, V.~A. Prisacariu, D.~Turmukhambetov, and E.~Brachmann, ``Map-free visual relocalization: Metric pose relative to a single image,'' in \emph{ECCV}, 2022.

\bibitem{shavit2021mstrans}
Y.~Shavit, R.~Ferens, and Y.~Keller, ``Learning multi-scene absolute pose regression with transformers,'' in \emph{2021 IEEE/CVF International Conference on Computer Vision (ICCV)}, 2021, pp. 2713--2722.

\bibitem{brahmbhatt2018geometry}
S.~Brahmbhatt, J.~Gu, K.~Kim, J.~Hays, and J.~Kautz, ``Geometry-aware learning of maps for camera localization,'' in \emph{CVPR}, 2018.

\bibitem{lee2023fusionloc}
J.~Lee, H.~Lee, and J.~Oh, ``Fusionloc: Camera-2d lidar fusion using multi-head self-attention for end-to-end serving robot relocalization,'' \emph{IEEE Access}, vol.~11, pp. 75\,121--75\,133, 2023.

\bibitem{radford2021learning}
A.~Radford, J.~W. Kim, C.~Hallacy, A.~Ramesh, G.~Goh, S.~Agarwal, G.~Sastry, A.~Askell, P.~Mishkin, J.~Clark \emph{et~al.}, ``Learning transferable visual models from natural language supervision,'' in \emph{International conference on machine learning}.\hskip 1em plus 0.5em minus 0.4em\relax PMLR, 2021, pp. 8748--8763.

\bibitem{gao2021clip}
P.~Gao, S.~Geng, R.~Zhang, T.~Ma, R.~Fang, Y.~Zhang, H.~Li, and Y.~Qiao, ``Clip-adapter: Better vision-language models with feature adapters,'' \emph{arXiv preprint arXiv:2110.04544}, 2021.

\bibitem{li2022envedit}
J.~Li, H.~Tan, and M.~Bansal, ``Envedit: Environment editing for vision-and-language navigation,'' in \emph{2022 IEEE/CVF Conference on Computer Vision and Pattern Recognition (CVPR)}, 2022, pp. 15\,386--15\,396.

\bibitem{zhong2021SGGfromNLS}
Y.~Zhong, J.~Shi, J.~Yang, C.~Xu, and Y.~Li, ``Learning to generate scene graph from natural language supervision,'' in \emph{ICCV}, 2021.

\bibitem{rao2021denseclip}
Y.~Rao, W.~Zhao, G.~Chen, Y.~Tang, Z.~Zhu, G.~Huang, J.~Zhou, and J.~Lu, ``Denseclip: Language-guided dense prediction with context-aware prompting,'' in \emph{Proceedings of the IEEE Conference on Computer Vision and Pattern Recognition (CVPR)}, 2022.

\bibitem{Mirjalili2023fmloc}
R.~Mirjalili, M.~Krawez, and W.~Burgard, ``Fm-loc: Using foundation models for improved vision-based localization,'' in \emph{2023 IEEE/RSJ International Conference on Intelligent Robots and Systems (IROS)}, 2023, pp. 1381--1387.

\bibitem{Matsuzaki2024CLIPLocML}
S.~Matsuzaki, T.~Sugino, K.~Tanaka, Z.~Sha, S.~Nakaoka, S.~Yoshizawa, and K.~Shintani, ``Clip-loc: Multi-modal landmark association for global localization in object-based maps,'' \emph{2024 IEEE International Conference on Robotics and Automation (ICRA)}, pp. 13\,673--13\,679, 2024.

\bibitem{manvi2024geollm}
R.~Manvi, S.~Khanna, G.~Mai, M.~Burke, D.~B. Lobell, and S.~Ermon, ``Geollm: Extracting geospatial knowledge from large language models,'' in \emph{The Twelfth International Conference on Learning Representations}, 2024.

\bibitem{li2024georeasoner}
L.~Li, Y.~Ye, B.~Jiang, and W.~Zeng, ``Georeasoner: Geo-localization with reasoning in street views using a large vision-language model,'' in \emph{International Conference on Machine Learning (ICML)}, 2024.

\bibitem{shen2023advancing}
F.~Shen, H.~Ye, J.~Zhang, C.~Wang, X.~Han, and W.~Yang, ``Advancing pose-guided image synthesis with progressive conditional diffusion models,'' \emph{arXiv preprint arXiv:2310.06313}, 2023.

\bibitem{shen2025boosting}
F.~Shen, H.~Ye, S.~Liu, J.~Zhang, C.~Wang, X.~Han, and Y.~Wei, ``Boosting consistency in story visualization with rich-contextual conditional diffusion models,'' in \emph{Proceedings of the AAAI Conference on Artificial Intelligence}, vol.~39, no.~7, 2025, pp. 6785--6794.

\bibitem{7scenes}
A.~Criminisi, J.~Shotton, B.~Glocker, S.~Izadi, and A.~Fitzgibbon, ``7-scenes dataset,'' 2013.

\bibitem{shavitferensirpnet}
Y.~Shavit and R.~Ferens, ``Do we really need scene-specific pose encoders?''\hskip 1em plus 0.5em minus 0.4em\relax IEEE, 2021, pp. 3186--3192.

\bibitem{mel2017hourglass}
I.~Melekhov, J.~Ylioinas, J.~Kannala, and E.~Rahtu, ``Image-based localization using hourglass networks,'' in \emph{2017 IEEE International Conference on Computer Vision Workshops (ICCVW)}, 2017, pp. 870--877.

\end{thebibliography}
%
%

\end{document}